\documentclass[10pt,twocolumn,letterpaper]{article}

\usepackage{cvpr}              % To produce the CAMERA-READY version
\definecolor{cvprblue}{rgb}{0.21,0.49,0.74}
\usepackage[pagebackref,breaklinks,colorlinks,allcolors=cvprblue]{hyperref}

\def\paperID{4355} % *** Enter the Paper ID here
\def\confName{CVPR}
\def\confYear{2026}

\title{Efficient Document Parsing via Parallel Token Prediction}

\author{
Lei Li$^{1}$\quad
Ze Zhao$^{1}$\quad
Meng Li$^{1,2}$\quad
Zhongwang Lun\quad
Yi Yuan$^{1}$\quad
Xingjing Lu$^{1}$\\
Zheng Wei$^{1\dag}$\quad
Jiang Bian$^{1}$\quad
Zang Li$^{1}$\\
{$^1$Platform and Content Group, Tencent\quad
$^2$Renmin University of China} \\
\texttt{arleyli@tencent.com, hemingwei@tencent.com}
}

\usepackage{makecell}
\usepackage{multirow}
\usepackage{amsmath}
\usepackage{wrapfig}
\usepackage{amssymb}
\usepackage{hyperref}
\usepackage{colortbl}

\begin{document}
\maketitle
\renewcommand{\thefootnote}{}
\footnotetext{$^\dag$ Corresponding author.}
\footnotetext{Code is available at \url{https://github.com/flow3rdown/PTP-OCR}.}
\renewcommand{\thefootnote}{\arabic{footnote}}

\begin{abstract}
Document parsing, as a fundamental yet crucial vision task, is being revolutionized by vision-language models (VLMs). However, the autoregressive (AR) decoding inherent to VLMs creates a significant bottleneck, severely limiting parsing speed. In this paper, we propose \textbf{Parallel-Token Prediction (PTP)}, a plugable, model-agnostic and simple-yet-effective method that enables VLMs to generate multiple future tokens in parallel with improved sample efficiency. Specifically, we insert some learnable tokens into the input sequence and design corresponding training objectives to equip the model with parallel decoding capabilities for document parsing. 
Furthermore, to support effective training, we develop a comprehensive data generation pipeline that efficiently produces large-scale, high-quality document parsing training data for VLMs. Extensive experiments on OmniDocBench and olmOCR-bench demonstrate that our method not only significantly improves decoding speed (1.6$\times$-2.2$\times$) but also reduces model hallucinations and exhibits strong generalization abilities.

\end{abstract} 

\section{Introduction}
\label{sec:intro}

% 1. document parsing任务的定义、重要性：基石任务：RAG、构建数据、理解数字世界
% 2. 基本方法：pipeline方法和基于VLM的end2end方法
% 3. VLM的致命缺陷：速度太慢，原因是因为输出的文字多加上NTP的特性
% 4. 联系一下CTC的方法 并行解码的特性， 但是很难直接用于VLM
% 5. 提出本文的思想，并行解码，定义、怎么做、带来的效益
% 6. 总结

Document parsing, also known as document content extraction~\cite{zhang2024document}, aims to transform unstructured or semi-structured documents into structured, machine-readable outputs. 
This process involves accurately identifying and reconstructing diverse elements including text, images, formulas, and tables while preserving their logical ordering and hierarchical relationships as presented in the original documents. 
As a cornerstone task in multimodal understanding, document parsing plays a critical role in enabling advanced applications such as Retrieval-Augmented Generation (RAG)~\cite{lin2024revolutionizing,zhang2025ocr}, document analysis~\cite{tang2023unifying,bai2022wukong}, and data management~\cite{qwen25vl, wei2025deepseek}, establishing a solid foundation for enabling machines to comprehend the digital world.
Early document parsing methods predominantly adopted pipeline-based approaches~\cite{cui2025paddleocr,wang2024mineruopensourcesolutionprecise,vik2024marker,poznanski2025olmocr}, which decomposed the task into sequential modules, suffering from error accumulation and limited end-to-end optimization. With recent advances in Vision-Language Models (VLMs), an increasing number of methods have begun leveraging VLMs to revolutionize the document parsing task, either through end-to-end generation~\cite{nougat,kim2021donut,wei2025deepseek,poznanski2025olmocr} or by integrating VLMs into specific pipeline stages~\cite{chai2024docfusion,dolphin,niu2025mineru25decoupledvisionlanguagemodel,liu2025points} for improved multi-element recognition.

% However, every coin has two sides. While VLMs have brought substantial performance improvements to document parsing, the autoregressive (AR) nature with next-token prediction (NTP) and the non-parallelizable characteristic of end-to-end parsing have introduced \textbf{parsing speed as an emerging challenge}.
% Several studies have begun exploring efficiency improvements for VLM-based parsing from various perspectives. MinerU2.5~\cite{niu2025mineru25decoupledvisionlanguagemodel} introduces an optimized table structure language to reduce output sequence length, while implementing fine-grained deployment optimizations to enhance overall throughput. Deepseek-OCR~\cite{wei2025deepseek} tackles the problem from the input side by proposing DeepEncoder, which substantially reduces the token count for high-resolution images, thereby decreasing the prefill latency of VLMs. LightOnOCR~\cite{lightonocr2025} employs a frequency-based pruning method to eliminate rarely-used or unused tokens from the vocabulary, thereby reducing unnecessary model capacity and computational overhead without compromising performance, ultimately improving overall efficiency. While the aforementioned works can improve overall parsing speed, they address the symptoms rather than the root cause. For VLMs, the generation phase constitutes the predominant latency bottleneck, particularly in document parsing tasks with dense text and information-rich content.

However, as a real-world application-oriented task, document parsing demands not only high accuracy but also efficient processing speed, particularly for large-scale deployment scenarios. While VLMs have achieved remarkable improvements in parsing quality, their inherent autoregressive (AR) generation mechanism with next-token prediction (NTP) introduces a singnificant efficiency bottleneck. Recent efforts have explored various optimization strategies to accelerate VLM-based parsing, including output sequence compression~\cite{niu2025mineru25decoupledvisionlanguagemodel}, visual token reduction~\cite{wei2025deepseek}, and model parameter pruning~\cite{lightonocr2025}. Despite these advances, \textbf{the sequential generation paradigm remains the inherent bottleneck}, as the autoregressive decoding process leading to substantial latency that grows proportionally with document complexity and content density.
% Considering that the essence of OCR tasks lies in accurate transcription rather than semantic understanding, we can naturally decompose an image into multiple patches and perform parallel content recognition. This raises a fundamental question: \textit{Can this parallel recognition capability be inherently embedded within the model architecture?}
% Based on this, we propose \textbf{Parallel Token Prediction (PTP)}, a method that fundamentally accelerates the generation speed by enabling the model to produce multiple tokens within a single decoding step through specially designed register tokens~\cite{Timoth2024Register,MuToR}.
% Specifically, we insert learnable special tokens in training sequences and constrain these tokens to predict future tokens based on their positions. During inference, we append $N$ special tokens to the input sequence (denoted as PTP-$N$), thereby achieving an $N$-fold inference acceleration.
Considering that the essence of OCR tasks lies in accurate transcription rather than semantic understanding, we can naturally decompose an image into multiple patches and perform parallel content recognition. This raises a natural question: \textit{Can this parallel recognition capability be inherently embedded within the model itself?}
To address this challenge, we propose \textbf{Parallel Token Prediction (PTP)}, a novel training and inference framework that breaks the sequential generation bottleneck by enabling models to produce multiple tokens per decoding step. Specifically, we insert some learnable register tokens~\cite{Timoth2024Register,MuToR} into training sequences and optimize them to predict future tokens based on their positions. During inference, by appending $N$ special tokens to the input sequence, the model generates $N$ tokens in parallel within each decoding step, achieving theoretical $N$-fold acceleration. Extensive experiments on the OmniDocBench~\cite{OmniDocBench} dataset validate that PTP delivers significant throughput improvements while preserving model accuracy: PTP-1 attains 1.6$\times$ throughput over the NTP baseline, and PTP-2 achieves 2.2$\times$ acceleration. Furthermore, we generalize PTP to broader vision-language understanding tasks and synergize it with speculative decoding~\cite{leviathan2023fast,NEURIPS2018_c4127b91}, resulting in an impressive 82\% acceptance ratio.

To summarize, our key contributions are encapsulated as follows:
(1) We propose Parallel Token Prediction (PTP), a model-agnostic, plugable, and highly efficient acceleration method for document parsing. PTP achieves 1.6$\times$-2.2$\times$ throughput improvements without compromising accuracy.
(2) We construct a high-quality layout-level document parsing dataset through an automated generation framework that integrates multiple types of VLMs for data annotation, coupled with sophisticated filtering and deduplication strategies to ensure data quality.
(3) We conduct comprehensive analyses and ablation studies, validating the effectiveness of PTP and further exploring its potential in vision-language understanding (VLU) scenarios.

\section{Related Work}
\label{sec:related_work}

\textbf{Document Parsing Approaches.}
Document parsing methods can be broadly categorized into two approaches:
\textbf{(i) Pipeline-based Approaches:} These methods~\cite{vik2024marker,wang2024mineruopensourcesolutionprecise} decompose document parsing into sequential modular tasks, including layout analysis, text, formula and table recognition, and reading order detection, etc. Each module employs a specialized model optimized for its specific task. While enabling fine-grained optimization and interpretability, they suffer from error accumulation across stages and exhibits degraded performance in challenging or domain-specific scenarios.
\textbf{(ii) VLM-based Approaches:} These methods leverage general or domain-specific vision-language models to replace multiple modular components, thereby simplifying the parsing pipeline.
Early works~\cite{nougat,wei2024vary,GOT} introduce end-to-end OCR-free VLMs that directly parse document images, eliminating error propagation, but remain constrained by scalability and efficiency concerns.
Recent approaches~\cite{dolphin,MonkeyOCR,niu2025mineru25decoupledvisionlanguagemodel} adopt a hybrid strategies combining layout analysis with VLM-based recognition. While this strategy effectively leverages both the efficiency of pipeline methods and the accuracy of VLMs, two critical limitations persist: (1) autoregressive decoding inherently limits parsing speed, and (2) the scarcity of large-scale, high-quality training data poses challenges for model development.

\textbf{Efficient Document Parsing.}
While autoregressive models improve OCR accuracy and robustness, efficiency remains a critical bottleneck. Existing acceleration approaches can be categorized as follows:
\textbf{(i) Multi-Token Prediction: }  Early works~\cite{(CASS-NAT,NARVL,cpdd} employ non-autoregressive (NAR) vision-language models trained with Connectionist Temporal Classification (CTC) loss to achieve multi-token prediction. However, these methods require complex architectural modifications, exhibit limited performance, and are restricted to span-level OCR tasks, failing to scale to paragraph- or document-level parsing. Recent efforts~\cite{mtp,liu2024deepseek} introduce auxiliary MTP heads to enable multi-token prediction in language models, but their applications to document parsing remain unexplored.
\textbf{(ii) Sequence Compression: }  Recent studies reduce computational cost by shortening input or output sequences. \cite{niu2025mineru25decoupledvisionlanguagemodel} designs compact representation languages for formulas and tables to reduce output tokens, thereby improving throughput. \cite{wei2025deepseek} proposes DeepEncoder to compress visual representations, reducing input tokens and accelerating prefill stage. \cite{lightonocr2025} prunes redundant vocabulary tokens to decrease model capacity and decoding overhead. While achieving moderate efficiency gains, these methods do not fundamentally address the autoregressive (AR) decoding bottleneck.
In this work, we propose Parallel Token Prediction (PTP), which enables parallel decoding in VLMs without sacrificing performance. PTP is model-agnostic and orthogonal to existing architectures and acceleration techniques, delivering substantial improvements in parsing efficiency.

\begin{figure*}[!t]
\centering
\includegraphics[scale=0.88]{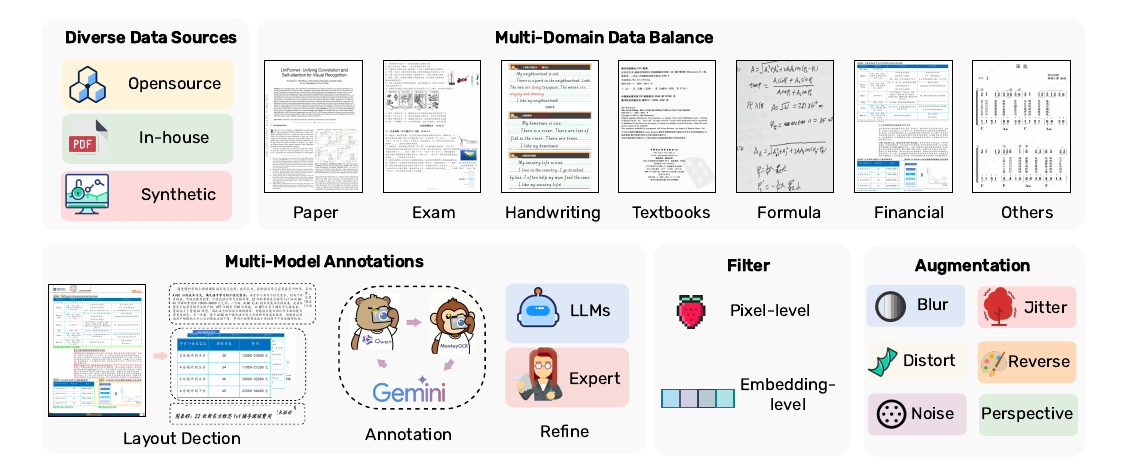}
\caption{An illustration of the data creation pipeline. }
\label{fig:dataset}
\end{figure*}

\section{Dataset Engine}
Current document parsing and OCR datasets mainly focus on span-level or file-level annotations, with a critical shortage of layout-level data. Moreover, existing datasets exhibit limited diversity in document types and difficulty levels, hindering model generalization to real-world scenarios.
To address these limitations, we develop a comprehensive and scalable data collection, annotation and cleaning pipeline, as shown in~\cref{fig:dataset}.

\subsection{Data Curation}
We begin by constructing a diverse document resource pool comprising 200k pages sourced through three channels: open-source datasets, in-house data, and synthetic generated data. 
We ensure that each document page is valid and contains parsable elements.
To maintain diversity and prevent category imbalance, we train a document classification and difficulty assessment model, which can identify document types (e.g. academic papers, technical reports, hand-writings) and difficulty levels, assisting us in controlling the distribution and achieving balanced representation across categories. More details in Supplementary Materials.

\subsection{Data Annotation}
We employ a layout analysis model~\cite{sun2025pp} to partition each document page into layout-based sub-regions (e.g., text paragraphs, tables, figures) to construct layout-level data.
To ensure the quality, we filter out sub-images that are too small, too large, or contain incomplete information due to boundary truncation.
Then we develop a multi-model collaborative annotation strategy that leverages three types of models, including a strong frontier VLM~\cite{comanici2025gemini}, an open-source VLM~\cite{qwen25vl}, and a specialized model~\cite{MonkeyOCR}.
Annotations from these models are aggregated through majority voting. The consolidated annotations are then refined via LLM-based post-processing to correct formatting errors, followed by selective manual review to ensure quality in cases with low confidence or high inter-model disagreement.

\subsection{Filtering and Statistics}
To ensure the quality and diversity of the final dataset, we implement a multi-stage filtering pipeline. We first remove corrupted images and samples with abnormal aspect ratios, which typically indicate scanning errors or improper cropping. To reduce redundancy and enhance diversity, we apply two complementary deduplication strategies: (i) Embedding-based similarity: We compute CLIP~\cite{radford2021learning} image embeddings and identify near-duplicates using cosine similarity to capture semantic-level redundancy; (ii) Perceptual hashing: We apply pHash with Hamming distance to detect visually similar images, capturing pixel-level similarity robust to minor transformations. 
Through this comprehensive filtering pipeline, 10\% of the collected data is removed, yielding a final dataset of 1.8M high-quality samples.

\section{Method}
\label{sec:method}

\begin{figure*}[!t]
\centering
\includegraphics[scale=0.85]{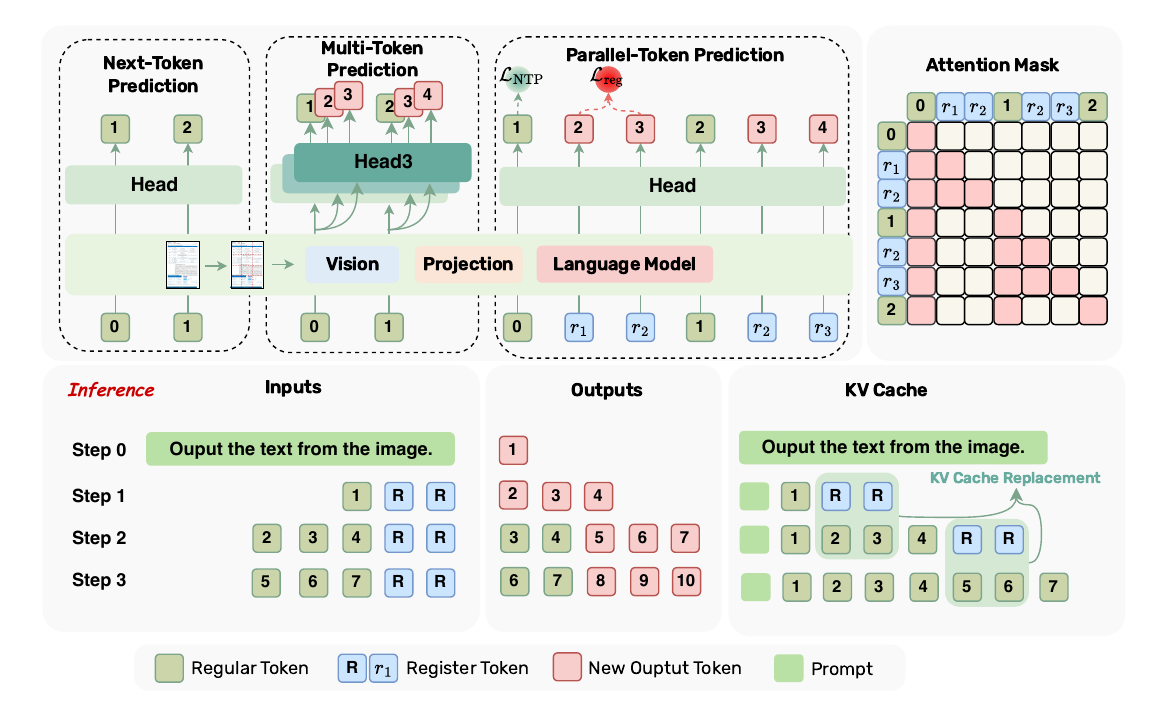}
\caption{Overview of our method. The upper illustrates the training architecture of our parallel-token prediction method, along with a visualization of the attention mask used during training. The lower depicts the inference process, where accelerated inference can be achieved by incorporating register tokens in each decoding step.}
\label{fig:method}
\end{figure*}

\subsection{Preliminaries}

\paragraph{Next-Token Prediction}
Next-Token Prediction (NTP) is the core objective of autoregressive vision-language models. Given a vision input $X_v$, a textual query $X_q$ and the answer $X_a$, NTP can be formulized as follows:
\begin{equation}
    P(x_1, \dots, x_l) = \prod_{i}^{l} P(x_{i+1} | X_v, X_q, X_{a,\leq i})
\end{equation}%
where $l$ is the length of answer $X_a$. For a model $P_\theta$ and dataset $\mathcal{D}$,  the training objective is to minimize the cross-entropy loss:
\begin{equation}
    \mathcal{L}_{\text{NTP}} = \mathbb{E}_{\mathcal{D}}[-\sum_i^l\log P_\theta (x_{i+1} | X_v, X_q, X_{a,\leq i})]
\end{equation}

\paragraph{Multi-Token Prediction}
\cite{mtp} proposed Multi-Token Prediction (MTP), which generalizes NTP by predicting multiple future tokens at once, as shown in \cref{fig:method}.
\begin{equation}
    \mathcal{L}_{\text{MTP}} = \mathbb{E}_{\mathcal{D}} [-\sum_i^l \log P_\theta (x_{i+1:i+n} | X_v, X_q, X_{a,\leq i} )]
\end{equation}%
where $n$ is the number of MTP heads.

\subsection{Parallel-Token Prediction}
\textbf{Overview.} Document parsing is essentially a high-certainty transcription task rather than an open-ended generation task, where the output is uniquely determined by the input image with minimal semantic ambiguity. 
Consider an image containing the text ``West Cowboy'': we can either process the entire image holistically or partition it into segments to separately recognize ``West'' and ``Cowboy'', both yielding identical results. This observation reveals an inherent parallelizability in document parsing that remains unexploited in previous works.
Building upon this insight, we propose Parallel Token Prediction (PTP), which enables models to simultaneously attend to and recognize multiple characters within an image, substantially improving generation efficiency. 
Specifically, following~\cite{Timoth2024Register,MuToR}, we introduce a set of learnable continuous tokens, termed \textit{registers}, appended after each token in the training sequence. Each register is trained to predict future tokens based on its relative distance from the preceding context. Through carefully designed training objectives, these registers acquire the capability to perform accurate multi-step-ahead predictions.

\textbf{Register Tokens.} \cite{Timoth2024Register} first introduced registers as additional learnable tokens appended to input sequences to store global information and absorb high-norm outlier features. Inspired by this, we repurpose registers for capturing features from distinct regions of the image and predict future tokens in parallel. Notably, all register tokens share the same token ID and learnable embedding, yet through contextual conditioning, they dynamically perform region-specific predictions at different positional offsets.

\textbf{Training.} Given $X_a=(x_1,x_2,\dots,x_{l-1}, x_l)$ as the answer token sequence to be trained, we insert continuous register tokens after each token (as shown in~\cref{fig:method}):
\begin{equation}
    \hat X_a=(x_1, [r_2, r_3], x_2, [r_3,r_4], \dots, x_{l-1}, [r_{l}, r_{l+1}], x_l)
\label{eq:input}
\end{equation}
where each regular token $x_i$ is augmented with $n$ subsequent continuous register tokens $r_j$ (here $n=2$). 
All register tokens share a single learnable embedding but differ in their positional encodings, enabling them to predict future tokens at position-dependent offsets. 
Specifically, $r_{i+1}$ placed immediately after $x_i$ is trained to predict $x_{i+2}$, while $r_{i+2}$ predicts $x_{i+3}$ and so on. Accordingly, the shifted training objective corresponding to Eq.~\ref{eq:input} becomes:
\begin{equation}
    \mathcal{O}_a=(x_2, [x_3, x_4], x_3, [x_4, x_5], \dots, x_l, [x_l, x_l])
\end{equation}

To ensure independent training between regular tokens $x_i$ and register tokens $r_i$, we modify the causal attention mask to enforce the following constraints: 
(1) Regular tokens attend only to preceding regular tokens and remain isolated from all register tokens. 
(2) Register tokens attend to all preceding regular tokens, as well as preceding register tokens within the same group (\emph{i.e.}, register tokens following the same regular token).
(3) Register tokens from different groups are mutually isolated and do not interact.

Since our method preserves the original model architecture, we adjust the position IDs of register tokens to enable accurate future token predictoin. Specifically, register token $r_i$ is assigned a position ID equal to its preceding regular token $x_{i-1}$ plus one. Similarly, register token $r_{i+1}$ receives a position ID one greater than $r_i$. Consequently, the position ID sequence corresponding to~\cref{eq:input} is:
\begin{equation}
    \mathcal{P}_a = (1, [2, 3], 2, [3, 4], \dots, l-1, [l, l+1], l)
\end{equation}
where we suppose the position id starts from 1.

During training, regular tokens are optimized using the standard NTP loss, while register tokens are optimized with the following loss:
\begin{equation}
    \mathcal{L}_{\text{reg}} = \mathbb{E}_\mathcal{D}[-\sum_i^l
    \sum_j^n\log P_\theta (x_{i+j+1}|X_{a, \le i}, r_{i+j})]
\end{equation}
Due to our meticulously crafted causal attention mask, regular tokens remain unaffected by register tokens throughout the training process.
Finally, the training loss of our PTP approach is defined as:
\begin{equation}
    \mathcal{L}_{\text{PTP}} = \alpha * \mathcal{L}_\text{NTP} + (1-\alpha)*\mathcal{L}_\text{reg}
\end{equation}
where $\alpha \in (0,1)$ controls relative weight of each loss term.

\subsection{Inference and Analysis}
\label{sec:inference}
Unlike~\cite{MuToR}, we do not discard register tokens during inference. Instead, we fully leverage their learned ability to predict future tokens for decoding acceleration. As illustrated in~\cref{fig:method}, at each decoding step, we append 
$n$ additional register tokens after the original input, enabling the model to generate $n+1$ new predictions per step. 
Subsequently, we can estimate the speedup ratio (SR) as follows:
\begin{equation}
    \text{SR} \approx \frac{(1+n)\times L_\theta}{L_\theta^\prime}
\end{equation}
where $L_\theta$ denotes the latency of the model per decode step. $L_\theta^\prime$ denotes the latency of a single forward pass processing multiple tokens simultaneously.
While this may vary slightly from $L_\theta$ due to the hardware, the difference remains negligible when computational resources are sufficient.

Since we only append register tokens at the end of the sequence, our approach fully conforms to the causal LM setting, requiring no modifications to attention masks or positions. 
The only necessary operation is removing the KV cache corresponding to register tokens after each decoding step. This is because we subsequently perform a forward pass with the tokens predicted by register tokens, which generates more accurate KV cache compared to the speculative register token predictions.
Although this approach introduces a slight computational overhead ($L_\theta$ vs. $L_\theta^\prime$), it does not impact overall throughput when computational resources are sufficient, since the decoding phase is memory-bound rather than compute-bound. The additional computation is effectively absorbed within memory access latency.

\section{Experiments}
\label{sec:experiments}

% 1. 实验设定
% 2. 实验结果：效果和性能
% 3. 分析实验：效率上（MTP）；效果上（NTP、幻觉实验）
% 4. 泛化实验：模型、数量、场景
% 5. 消融实验

% Document Parsing: Omnibench
\begin{table*}[!t]
  \small
  \centering
  \resizebox{1\textwidth}{!}{
    \begin{tabular}{c|l|cccccccccc}
    \toprule
  \textbf{Model Type} & \textbf{Models} & \textbf{Slides} & \makecell{\textbf{Academic}\\\textbf{Papers}} & \textbf{Book} & \textbf{Textbook} & \makecell{\textbf{Exam}\\\textbf{Papers}} & \textbf{Magazine} & \textbf{News} & \textbf{Notes} & \makecell{\textbf{Financial}\\\textbf{Report}} & \textbf{Overall($\downarrow$)} \\
  % ======================== table head ========================
    \midrule
    \multirow{3}{*}{\makecell{\textbf{Pipeline}\\\textbf{Tools}}}  
    & Marker-1.8.2~\cite{vik2024marker} & 0.1796 & 0.0412 & 0.1010 & 0.2908 & 0.2958 & 0.1111 & 0.2717 & 0.4656 & 0.0341 & 0.1990 \\
    & MinerU2-pipeline~\cite{wang2024mineruopensourcesolutionprecise} & 0.4244 & 0.0230 & 0.2628 & 0.1224 & 0.0822 & 0.3950 & 0.0736 & 0.2603 & 0.0411 & 0.1872 \\
    & PP-StructureV3~\cite{cui2025paddleocr}  & 0.0794 & 0.0236 & 0.0415 & 0.1107 & 0.0945 & 0.0722 & 0.0617 & 0.1236 & 0.0181 & 0.0695 \\
    
    \midrule
    \multirow{5}{*}{\makecell{\textbf{General}\\\textbf{VLMs}}} 
    & GPT-4o~\cite{achiam2023gpt}  & 0.1019 & 0.1203 & 0.1288 & 0.1599 & 0.1939 & 0.1420 & 0.6254 & 0.2611 & 0.3343 & 0.2297 \\
    & Gemini-2.5 Pro~\cite{comanici2025gemini}  & 0.0326 & \textbf{0.0182} & 0.0694 & 0.1618 & 0.0937 & 0.0161 & 0.1347 & 0.1169 & 0.0169 & 0.0734\\ 
    & InternVL3-76B~\cite{internvl3}  & 0.0349 & 0.1052 & 0.0629 & 0.0827 & 0.1007 & 0.0406 & 0.5826 & \textbf{0.0924} & 0.0665 & 0.1298 \\
    & Qwen2.5-VL-3B~\cite{qwen25vl} & 0.1809 & 0.1489 & 0.1895 & 0.2607 & 0.3527 & 0.1599 & 0.1690 & 0.2237 & 0.1893 & 0.2083  \\
    & Qwen2.5-VL-72B~\cite{qwen25vl}  & 0.0422 & 0.0801 & 0.0586 & 0.1146 & \underline{0.0681} & 0.0964 & 0.2380 & 0.1232 & 0.0264 & 0.0924\\
    
    \midrule
    \multirow{7}{*}{\makecell{\textbf{Specialized}\\\textbf{VLMs}}}
    & Dolphin~\cite{dolphin}  & 0.0957 & 0.0453 & 0.0616 & 0.1333 & 0.1684 & 0.0702 & 0.2388 & 0.2561 & 0.0186 & 0.1209 \\
    & olmOCR-7B~\cite{olmOCR}  & 0.0497 & 0.0365 & 0.0539 & 0.1204 & 0.0728 & 0.0697 & 0.2916 & 0.1220 & 0.0459 & 0.0957 \\
    & MonkeyOCR-pro-1.2B~\cite{MonkeyOCR}  & 0.0961 & 0.0354 & 0.0530 & 0.1110 & 0.0887 & 0.0494 & 0.0995 & 0.1686 & 0.0198 & 0.0802 \\ 
    & MonkeyOCR-3B~\cite{MonkeyOCR}  & 0.0904 & 0.0362 & 0.0489 & 0.1072 & 0.0745 & 0.0475 & 0.0962 & 0.1165 & 0.0196 & 0.0708 \\
    & dots.ocr~\cite{dots.ocr} & \textbf{0.0290} & 0.0231 & 0.0433 & 0.0788 & \textbf{0.0467} & 0.0221 & 0.0667 & 0.1116 & \underline{0.0076} & 0.0477 \\
    & MinerU2.5~\cite{niu2025mineru25decoupledvisionlanguagemodel}  & \underline{0.0294} & 0.0235 & \underline{0.0332} & \textbf{0.0499} & 0.0681 & 0.0316 & 0.0540 & 0.1161 & \underline{0.0104} &  \underline{0.0462} \\
    & Qwen2.5-VL-3B-NTP & 0.0812 & 0.0273 & 0.0460 & 0.0835 & 0.0969 & 0.0236 &  \underline{0.0366} &  \underline{0.0982} & 0.0329 & 0.0585 \\
     
    \midrule   
    \rowcolor{blue!10}
     & 
     Qwen2.5-VL-3B-PTP0 & 0.0744 & 0.0327 & 0.0427 & 0.0409 & 0.0797 & 0.0242 & 0.0404 & 0.0950 & 0.0316 & 0.0513 \\
    \rowcolor{blue!10}
    \textbf{Ours}
    & 
    Qwen2.5-VL-3B-PTP1 & 0.0572 & \underline{0.0213} & \textbf{0.0176} & \underline{0.0631} & 0.0779 & \textbf{0.0079} & \textbf{0.0392} & 0.0994 & \textbf{0.0047} & \textbf{0.0431}  \\
    \rowcolor{blue!10}
    & 
    Qwen2.5-VL-3B-PTP2 & 0.0616 & 0.0351 & \underline{0.0203} & 0.0853 & 0.0992 & \underline{0.0122} & 0.0456 & 0.1627 & 0.0083 & 0.0589 \\
         
    \bottomrule
    \end{tabular}%  
  }
    \caption{Document Parsing Performance in Text Edit Distance on OmniDocBench: evaluation using edit distance across 9 PDF page types.}    
  \label{tab:OmniDocbench-results-2}%
\end{table*}%

%olmBench
\begin{table*}[!t]
\centering
\small
\begin{tabular}{lccccccccc}
\toprule
\textbf{Model} & \textbf{Overall($\uparrow$)} & \textbf{AR} & \textbf{OSM} & \textbf{TA} & \textbf{OS} & \textbf{HF} & \textbf{MC} & \textbf{LTT} & \textbf{Base}  \\
\midrule
    MinerU2-pipeline~\cite{wang2024mineruopensourcesolutionprecise} & 55.6 & 61.8 & 13.5 & 60.9 & 17.3 & \textbf{96.6} & 59.0 & 39.1 & 96.6  \\
    GPT-4o~\cite{achiam2023gpt} & 63.2 & 44.1 & 37.6 & 69.1 & 40.9 & 94.2 & 68.9 & 54.1 & 96.7  \\
    Qwen2.5-VL-72B~\cite{qwen25vl} & 64.8 & \underline{72.2} & \underline{51.1} & 67.3 & 38.6 & 73.6 & 68.3 & 49.1 & 98.3  \\ 
    MonkeyOCR-pro-3B~\cite{MonkeyOCR} & 68.8 & 67.7 & 28.4 & 74.6 & 36.1 & 91.2 & 76.6 & 80.1 & 95.3  \\
    olmOCR~\cite{olmOCR} & 71.8 & 63.9 & 41.0 & 72.9 & \textbf{43.9} & 95.1 & 77.3 & 81.2 & 98.9  \\
    dots.ocr~\cite{dots.ocr} & \underline{73.6} & 66.3 & 35.8 & \textbf{88.3} & \underline{40.9} & 94.1 & \textbf{82.4} & 81.2 & \textbf{99.5} \\
    MinerU2.5~\cite{niu2025mineru25decoupledvisionlanguagemodel} & \textbf{75.2} & \textbf{76.6} & \textbf{54.6} & \underline{84.9} & 33.7 & \textbf{96.6} & 78.2 & 83.5 & 93.7  \\
\midrule
    \rowcolor{blue!10}
    Qwen2.5-VL-3B-PTP-0 & 71.0 & 63.5 & 38.6 & 71.3 & 34.2 & 95.3 & 78.3 & \textbf{87.6} & \textbf{99.5} \\
    \rowcolor{blue!10}
    Qwen2.5-VL-3B-PTP-1 & 70.6 & 62.6 & 38.6 & 70.4 & 33.6 & \underline{96.2} & \underline{79.4} & \underline{84.7} & \underline{99.4}  \\
    \bottomrule
    \end{tabular}
\caption{Evaluation results on olmOCR-bench grouped by document types, including arXiv Math(AR), Old Scans Math (OSM), Tables (TA), Old Scans (OS), Headers Footers (HF), Multi Column (MC) and Long Tiny Text (LTT). Some results are sourced from the official reports of olmOCR-bench~\cite{olmOCR} and dots.ocr~\cite{dots.ocr}. The Overall Score (Overall) represents the average across all document types.}
\label{tab:olmocr-bench-results}
\end{table*}

% Document Parsing: Formula
\begin{table}[!t]
  \small
  \centering
  \resizebox{0.46\textwidth}{!}{
    \begin{tabular}{l|ccc}
    \toprule
    \textbf{Model} & \textbf{CDM($\uparrow$)} & \textbf{BLUE($\uparrow$)} & \textbf{Norm Edit($\downarrow$)} \\
  % ======================== table head ========================
    \midrule
    Mathpix~\cite{Mathpix2025} & 86.60 & \textbf{66.56} & 0.322 \\
    Pix2Tex & 73.90 & 46.00 & 0.337 \\
    UniMERNet-B~\cite{wang2024unimernet} & 85.00 & 60.84 & 0.238 \\
    GPT4o~\cite{achiam2023gpt} & 86.80 & 45.17 & 0.282  \\
    Qwen2.5-VL-3B~\cite{qwen25vl} & 24.11 & 53.59 & 0.331\\
    Qwen2.5-VL-3B-NTP & 71.65 & \underline{63.05} & \textbf{0.226} \\
    \midrule
    \rowcolor{blue!10}
    Qwen2.5-VL-3B-PTP0 & \textbf{91.59} & 63.38 & \underline{0.231} \\
    \rowcolor{blue!10}
    Qwen2.5-VL-3B-PTP1 & \underline{89.63} & 62.32 & 0.236 \\
    \rowcolor{blue!10}
    Qwen2.5-VL-3B-PTP2 & 77.23 & 57.92 & 0.284 \\
         
    \bottomrule
    \end{tabular}%  
  }
    \caption{Formula recognition results on OmniDocBench. We report results in terms of both CDM~\cite{wang2024unimernet}, BLEU and Edit Distance.}    
  \label{tab:formula}%
\end{table}%

\subsection{Experimental Settings}

\textbf{Datasets \& Baselines.} 
We primarily evaluate our method on OmniDocBench~\cite{OmniDocBench} and olmOCR-bench~\cite{olmOCR} document parsing benchmarks, focusing on text recognition and formula recognition performance. 
OmniDocBench is currently the most widely adopted benchmark for document parsing, designed to assess diverse document understanding in real-world scenarios. It encompasses nine document types, four layout types, and three language types, providing comprehensive coverage of practical document parsing challenges. 
olmOCR-bench comprises 1,402 PDF documents sourced from various repositories, organized into seven subsets.
We mainly compare with three types of methods: pipeline tools~\cite{vik2024marker,wang2024mineruopensourcesolutionprecise,cui2025paddleocr}, general VLMs~\cite{achiam2023gpt,comanici2025gemini,internvl3,qwen25vl} and specialized VLMs~\cite{dolphin,olmOCR,MonkeyOCR,dots.ocr,niu2025mineru25decoupledvisionlanguagemodel}.

\textbf{Implementation Details.}
Taking into account both performance and effectiveness, we employ the Qwen2.5-VL-3B-Instruct models as our base model and fine-tune it on our constructed dataset. During fine-tuning, we set the max number of register tokens to $n=3$ and the loss weight to $\alpha=0.5$. 
All experiments are conducted on 8 × A100 40GB GPUs for 1 epoch with a learning rate of $2e-5$.
All experiments are trained for 1 epoch with a learning rate of $2e-5$.
We freeze the vision encoder and aligner parameters, updating only the LLM weights. 
In all experiments, we denote models trained solely with $\mathcal{L}_\text{NTP}$ as $*$-NTP, and models trained with $\mathcal{L}_\text{PTP}$ as $*$-PTP-$n$, where $n$ indicates the number of inserted register tokens during inference.

\subsection{Main Results}

\textbf{PTP Enhances Recognition Accuracy.}
The main performance results of text recognition and formula recognition for all models are shown in~\cref{tab:OmniDocbench-results-2}, \cref{tab:olmocr-bench-results} and \cref{tab:formula}, respectively. 
Firstly, models fine-tuned on our constructed dataset achieve significant performance gains, matching or exceeding many specialized models while using substantially less training data (PTP-0 and NTP). Secondly, when incorporating one register token for parallel inference (PTP-1), the text recognition \textbf{performance not only remains intact but further improves}, surpassing other competing methods. 
This improvement may be attributed to PTP encouraging the model to better leverage surrounding contextual information, thereby reducing hallucinations and producing more accurate predictions.
Moreover, although formula recognition involves complex LaTeX syntax reasoning, PTP-1 achieves performance comparable to NTP while significantly accelerating inference.

\textbf{PTP Improves Throughput.}
We integrate the PTP implementation into KsanaLLM~\cite{tencent2025ksanallm} and evaluate the efficiency of PTP using an H20 (90G) GPU.
% We did not implement PTP in vLLM~\cite{kwon2023efficient} or SGLang~\cite{zheng2024sglang} due to their highly encapsulated architectures, which pose significant challenges for modification.
The results are presented in~\cref{fig:latency}, we observe that PTP effectively reduces both time per output token (TPOT) and average latency while significant improving decoding throughput. Specifically, \textbf{PTP-1 achieves 1.6× speedup over NTP, while PTP-2 attains 2.2× speedup.}

\subsection{Analysis}

\begin{figure*}[!t]
\centering
% \vskip -3mm
\includegraphics[scale=0.71]{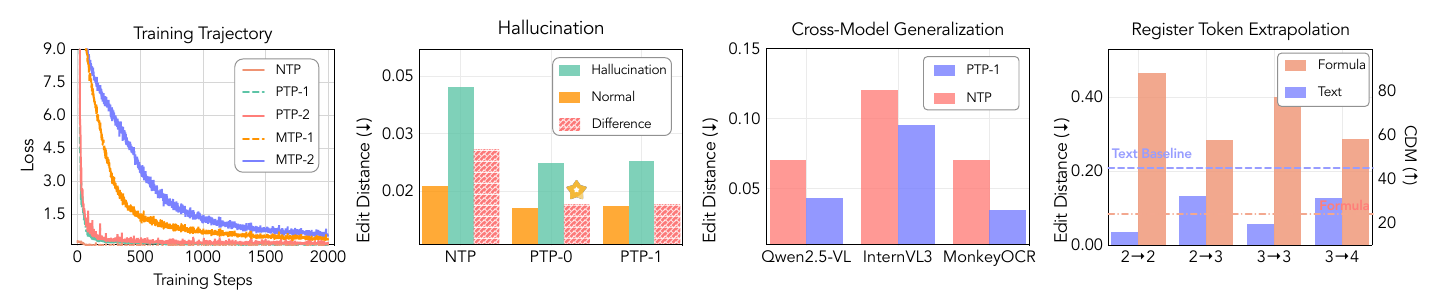}
\caption{\textbf{Left}: Training trajectories of NTP, PTP, and MTP; \textbf{Second}: Performance of different methods on normal vs. hallucination-prone data; \textbf{Third}: PTP performance across different model architectures; \textbf{Right}: Register token extrapolation results.}
\label{fig:analysis}
% \vskip -6mm
\end{figure*}

\textbf{Efficiency Analysis. }
To comprehensively evaluate the efficiency of our proposed PTP method, we conduct comparative analysis from both training and inference perspectives against NTP and MTP approaches. For fair comparison, we follow the MTP architecture from Mimo~\cite{Mimo} and adopt the training strategy from FastMTP~\cite{cai2025fastmtp} to augment Qwen2.5-VL with shared MTP heads and blocks. All models are fine-tuned on identical datasets with same training setting. \textit{(i) Training Efficiency.}
The training trajectories in~\cref{fig:analysis} reveal significant efficiency advantages of PTP over MTP. While both methods exhibit initially high loss values, \textbf{PTP demonstrates rapid loss reduction and achieves fast convergence}, whereas MTP requires substantially more training steps to reach comparable performance. Notably, PTP achieves loss levels on par with NTP while substantially outperforming MTP. Additionally, PTP exhibits consistent convergence patterns across different configurations (PTP-1 and PTP-2), while MTP shows notable sensitivity to the number of prediction heads, with MTP-2  exhibiting significantly slower convergence.
This maybe attribute to MTP introducing additional head and block parameters, whereas PTP requires only learnable register tokens without architectural modifications, resulting in superior training efficiency and stability.
\begin{figure*}[!t]
\centering
% \vskip -3mm
\includegraphics[scale=0.7]{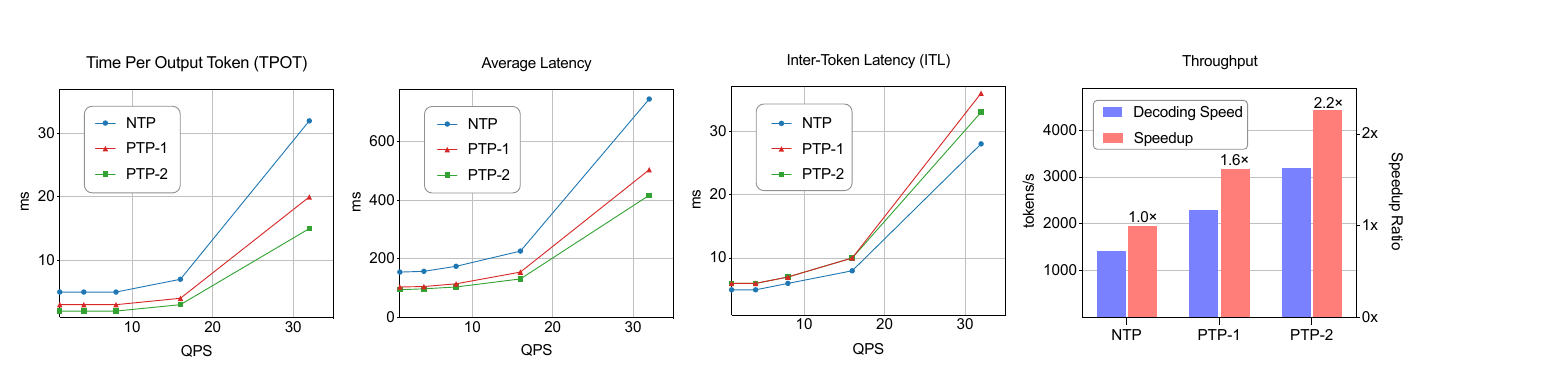}
\caption{Performance comparison between NTP and PTP, including average TPOT, ITL, and latency under different QPS levels, as well as decoding speed and speedup ratio in synchronous mode, which are measured on OmniDocBench 16,886 images using an H20 GPU.
}
\label{fig:latency}
% \vskip -6mm
\end{figure*}%
\textit{(ii) Inference Efficiency. }
Our PTP method also demonstrates significant advantages during inference. As illustrated in~\cref{fig:latency}, PTP achieves substantial decoding acceleration compared to NTP. While MTP employs a self-speculative approach that yields results comparable to NTP (but underperforms PTP-1), it achieves only a 70\% acceptance rate, resulting in lower speedup than PTP (the acceptance rate of PTP is 100\% theoretically).
The superior performance of PTP stems from its architectural advantages: PTP achieves true parallelism across all predicted tokens through register tokens, whereas MTP introduces sequential dependencies between prediction heads, limiting parallelization efficiency. During inference, register tokens in PTP are processed through the complete model architecture identically to regular tokens, ensuring consistent representation quality and robust predictions.

\textbf{Exploring the Role of PTP.}
Our PTP method not only significantly improves inference efficiency but also positively impacts model performance on document parsing task.
\textit{(i) PTP Enhances NTP Performance.} 
Similar to observations in multi-token prediction approaches~\cite{wei2025deepseek,Mimo}, PTP densifies training signals by predicting multiple future tokens simultaneously, improving data efficiency and encouraging the model to develop superior long-term planning strategies. As shown in~\cref{tab:OmniDocbench-results-2}, PTP-0 trained with $\mathcal{L}_{\text{PTP}}$ consistently outperforms standard NTP trained with $\mathcal{L}_{\text{NTP}}$ across all benchmarks. Notably, both models use identical training data and configurations, with the only difference being the additional regularization objective $\mathcal{L}_{\text{reg}}$ in PTP training. This demonstrates that our parallel prediction framework not only accelerates inference but also improves the model's representational capabilities.
\textit{(ii) Hallucination Mitigation.} Hallucination poses a critical challenge for VLMs in OCR tasks, particularly when processing images with blurred, distorted, or linguistically irregular text. Traditional autoregressive generation via NTP is inherently susceptible to error propagation, as models rely heavily on preceding context and may generate erroneous tokens through auto-completion or over-correction mechanisms~\cite{he2025seeingbelievingmitigatingocr}.
In contrast, PTP's parallel generation mechanism enables simultaneous attention to multiple image regions while generating corresponding tokens concurrently, thereby reducing over-dependence on sequential context. To quantify this advantage, we construct a controlled hallucination benchmark by systematically injecting noise into ground-truth annotations and their corresponding images through random word replacement and deletion (details in Supplementary Materials). As illustrated in~\cref{fig:analysis}.Second, PTP exhibits substantially lower hallucination rates compared to NTP. This improvement stems from PTP's ability to leverage global visual information more effectively, making predictions based on direct visual evidence rather than potentially corrupted contextual cues.

% 3. Generalization analysis - model & parallel
% \subsection{Generalization Analysis.}
% To validate the generalizability of our method, we conduct generalization studies from two perspectives:
% (1) \textbf{Cross-Model Generalization}: We evaluate PTP across models spanning different architectural paradigms, scales, and design principles. As shown in~\cref{fig:analysis}, PTP yields consistent performance gains across all configurations, validating its model-agnostic nature, plug-and-play compatibility, and minimal architectural requirements.
% (2) \textbf{Register Token Extrapolation}: We investigate scenarios where the number of register tokens differs between training and inference. Specifically, we evaluate configurations with $n=2$ tokens during training and $n=3$ during inference (PTP-2 $\rightarrow$ PTP-3), as well as $n=3$ tokens during training and $n=4$ during inference (PTP-3 $\rightarrow$ PTP-4). While some performance degradation is observed, the results still surpass the naive baseline model. Moreover, this performance gap can potentially be mitigated by combining with speculative decoding (described in~\cref{sec:spec}).

\subsection{Generalizability Study}
To comprehensively validate the generalizability of our PTP method, we conduct systematic evaluations across three critical dimensions: cross-model generalization, register token extrapolation, and cross-task adaptation.

\textbf{Cross-Model Generalization.} 
We evaluate PTP across models spanning different architectural paradigms, scales, and domains to assess its model-agnostic nature. As shown in~\cref{fig:analysis}.Third, PTP consistently yields substantial performance gains across all tested configurations, from compact models to large-scale architectures. This demonstrates PTP's flexible compatibility and minimal architectural requirements, making it broadly applicable to diverse vision-language models without extensive customization.

\textbf{Register Token Extrapolation.} 
We investigate the robustness of PTP when the number of register tokens differs between training and inference phases, as shown in~\cref{fig:analysis}.Right. Specifically, we examine two extrapolation scenarios: training with $n=2$  tokens while inference with $n=3$  (PTP-2 $\rightarrow$ PTP-3), and training with $n=3$ tokens while inferring with $n=4$ (PTP-3 $\rightarrow$ PTP-4). Although modest performance degradation is observed under these mismatched conditions, the results consistently surpass the vanilla model without register tokens. This degradation can be attributed to distributional shifts in learned token representations and can be effectively mitigated through self-speculative decoding mechanisms (discussed below).

\textbf{Task Domain Generalization.} Beyond document parsing task, we further investigate whether PTP generalizes to Vision-Language Understanding (VLU) tasks that require complex reasoning. We select the ScienceQA~\cite{lu2022learn} benchmark, which is particularly challenging as answers incorporate explicit chain-of-thought (CoT) reasoning, making it particularly suitable for evaluating both accuracy and acceleration efficiency. We train on the official training split and report results on the test set. As shown in~\cref{tab:spec}, PTP-1 achieves comparable performance to standard NTP while delivering substantial latency reductions. 
Furthermore, our PTP can be seamlessly integrated with \textit{self-speculative decoding}, which incorporates verification mechanisms without requiring additional draft models (details in Supplementary Materials), achieving performance identical to NTP with an 82\% acceptance rate and minimal latency overhead. These findings substantiate the broad applicability of our method across diverse tasks.

\begin{table}[!t]
    \footnotesize
    \centering
    \scalebox{1.0}{
    \begin{tabular}{l|cc}
        \toprule
        \textbf{Model} & \textbf{Accuracy} & \textbf{Accept Rate (\%)} \\
        \midrule
        Qwen2.5-VL-3B-NTP & 92.21 & N/A \\
        Qwen2.5-VL-3B-PTP-1 & 91.72 & \cellcolor{red!10} 100 \\
        \quad w/ speculative decoding & \cellcolor{red!10} 92.21 & 82 \\
        \bottomrule
    \end{tabular}
    }
    \caption{Results on ScienceQA dataset.}
    \label{tab:spec}
\end{table}

\begin{table}[!t]
  \small
  \centering
  \scalebox{0.80}{
    \begin{tabular}{l|ccc|c}
    \toprule
    \multirow{2}{*}{\textbf{Model}} & \multicolumn{3}{c|}{\textbf{Formula}} & \multirow{2}{*}{\textbf{Text Edit (↓)}} \\
    \cmidrule{2-4}
    & \textbf{CDM(↑)} & \textbf{BLEU(↑)} & \textbf{Edit(↓)} & \\
    \midrule
  % ======================== table head ========================
    Qwen2.5-VL-3B & 24.11 & 53.59 & 0.331 & 0.208 \\
    Qwen2.5-VL-3B-PTP-1 & \cellcolor{red!10} 89.63 & \cellcolor{red!10} 62.32 & \cellcolor{red!10} 0.236 & \cellcolor{red!10} 0.043 \\
    \quad w/ distinct reg. & 89.24 & 61.78 & 0.247 & 0.052 \\
    \quad w/ interleaved reg. & 88.46 & 61.45 & 0.244 & 0.128 \\
    \quad w/o KV Cache & 39.43 & 35.96 & 0.476 & 0.070 \\
    \bottomrule
    \end{tabular}%  
  }
    \caption{Ablation study on text and formula recognition in OmniDocBench.}    
  \label{tab:ablation}%
\end{table}%

\subsection{Ablation Study}

\textbf{Shared vs. Distinct Register Embedding.} 
We investigate whether using a single shared learnable embedding across all positions of register outperforms position-specific distinct embeddings. As shown in~\cref{tab:ablation}, the shared embedding configuration achieves marginally better performance, consistent with observations in~\cite{MuToR}.

\textbf{Interleaved vs. Continuous Register Tokens.} 
We compare an interleaved insertion strategy based on~\cite{MuToR}, which inserts a single register token between regular tokens to predict future tokens at specified offsets, against our continuous approach that inserts a fixed number of sequential register tokens for incremental multi-token prediction. 
Experiments demonstrate clear advantages of our method, particularly for PTP-2. Intuitively, the continuous approach enables PTP-2 to leverage contextual information from PTP-1 during prediction, whereas the interleaved one predicts in isolation without access to intermediate information.

\textbf{Inference without KV Cache Replacement. } 
As detailed in~\cref{sec:inference}, register tokens serve purely as computational intermediates without carrying semantic content explicitly.  
Consequently, during inference, we discard all the KV cache entries associated with registers and replace cache states using their corresponding predicted tokens in the next decode step. 
Notably, this operation introduces only negligible additional computation without impacting inference latency when computation resources are sufficient.
Experimental results validate that the cache replacement mechanism is essential for maintaining performance.

% \subsection{Repurposing PTP for Speculative Decoding}
% \label{sec:spec}

% In this section, we further investigate whether parallel token prediction is effective for Vision-Language Understanding (VLU)  tasks. 
% We select the ScienceQA~\cite{lu2022learn} dataset, whose answers incorporate explicit chain-of-thought (CoT) reasoning, making it particularly suitable for evaluating acceleration efficiency.
% We train on the training set and evaluate on the test set. 
% As shown in~\cref{tab:spec}, compared to the NTP approach, PTP-1 exhibits only marginal performance degradation while substantially reducing latency. 
% Furthermore, PTP can be combined with speculative decoding to incorporate a verification mechanism, achieving performance identical to NTP with 82\% acceptance rate and modest latency overhead. 
% These findings substantiate the broad applicability of our method across diverse tasks.

\section{Conclusion}
\label{sec:conclusion}
In this paper, we propose PTP, a parallel token prediction method that enables VLMs to efficiently accelerate document parsing. Our contributions include: (1) a high-quality layout-level document parsing data generation framework, and (2) an architecture-agnostic, plugable and simple-yet-effective framework implementing parallel token prediction with registers injection. Experimental results demonstrate that our approach achieves 1.6$\times$-2.2$\times$ decoding speedup while fully preserving parsing accuracy.

{
    \small
    \bibliographystyle{ieeenat_fullname}
    \bibliography{main}
}

\clearpage
\setcounter{page}{1}
\maketitlesupplementary

% 附录
% 数据方面
% 数据来源、数据统计（表，包括类型、数量等）
% 数据标注、过滤和增强的细节

% 方法
% 如何与投机解码结合

% 实验方面：
% 实验设置的表格
% 速度方面：
    % 测试的细节
    % 计算量增加带来的影响
% register token的个数分析
% 幻觉测试的细节
% case的分析

\section{Details of Dataset}

\subsection{Document Resource Pool}

We collect raw document and layout-level OCR data from three channels:
\begin{itemize}
    \item \textbf{Open-source datasets:} We aggregate diverse datasets including layout analysis (DocIIENet\cite{xing2024dochienet}), handwriting (GNHK\cite{lee2021gnhk}, CASIA-HWDB2\cite{liu2011casia}), and mathematical formulas (Unimernet-1M\cite{wang2024unimernet}, HME100K\cite{yuan2022syntax}). These primarily contain page-level or layout-level images. After format standardization and normalization, we obtain nearly 200K image-text pairs.
    \item \textbf{In-house dataset:} Sourced from our internal document collections, featuring complex layouts and diverse document types. All sensitive and personally identifiable information has been rigorously filtered.
    \item \textbf{Synthetic dataset:} To address specific scenarios such as handwriting recognition, we render images with varying CSS styles, fonts, and corpus, improving model robustness on challenging cases.
\end{itemize}

\subsection{Data Annotation, Cleaning and Augmentation}

Following layout-based segmentation of raw documents, we perform comprehensive data annotation, cleaning, and augmentation:
\begin{itemize}
\item \textbf{Annotation.} We employ a multi-model annotation pipeline: Qwen2.5-VL-72B and MonkeyOCR-Pro-3B generate initial annotations, with confidence scores computed via edit distance between predictions. Low-confidence samples undergo refinement with Gemini-2.5-Pro. All annotations are standardized through rule-based post-processing, followed by LLM-assisted and manual verification for quality assurance.

\item \textbf{Cleaning.} We implement a multi-stage filtering process: (1) removal of corrupted images and samples with abnormal aspect ratios; (2) filtering of extremely low-confidence annotations; (3) duplicate detection via CLIP embedding similarity (cosine distance threshold) and perceptual hashing (pHash) for pixel-level redundancy removal.

\item \textbf{Augmentation.} We apply stochastic augmentations during training, including blur, color jitter, geometric distortion, horizontal flipping, Gaussian noise injection, and perspective transformation, to improve model robustness and generalization.
\end{itemize}

\section{Combining PTP with Speculative Decoding}
\begin{figure}[!h]
    \centering
    \includegraphics[width=0.46\textwidth]{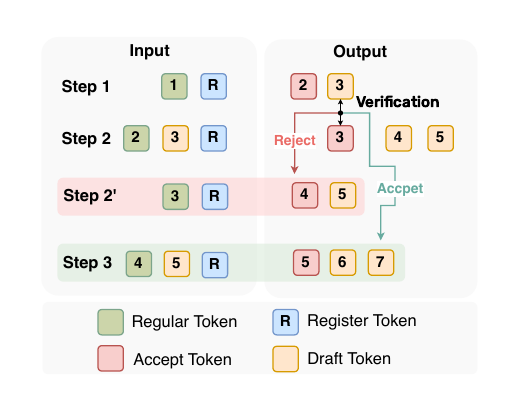}
    \caption{Speculative Decoding}
    \label{fig:spec}
\end{figure}

\begin{table}[!t]
    \footnotesize
    \centering
    \scalebox{0.9}{
    \begin{tabular}{l|c|ccc}
        \toprule
        \textbf{Dataset} & \textbf{Task} & \textbf{NTP} & \textbf{MTP} & \textbf{PTP-1} \\
        \midrule
        OmnidocBench & OCR & 0.0431 & 83\% & 95\% \\
        ScienceQA & VLU & 92.2 & 76\% & 82\% \\
        GSM8K & Math & 73.3 & 70\% & 88\% \\
        \bottomrule
    \end{tabular}
    }
    \caption{The performance of NTP, and acceptance ratios of MTP / PTP on various tasks.}
    \label{tab:re_ptp}
\end{table}

For tasks beyond image-to-text transcription, such as visual-language understanding (VLU), PTP may result in performance degradation relative to standard NTP.
However, our PTP method integrates seamlessly with speculative decoding through a self-verification mechanism for register token predictions, ensuring output consistency with standard NTP.
As shown in \cref{fig:spec}, we exploit the model's inherent predictions to validate register tokens from the previous decoding step, eliminating the need for external draft models. 
Specifically, tokens predicted by register tokens serve as draft candidates, which are subsequently verified against regular token predictions in the current step. Only verified tokens are retained in the final output sequence.
This self-verification approach is parameter-free, requiring no draft models or auxiliary layers. Compared to standard PTP, it introduces zero computational overhead when draft tokens are accepted, incurring additional cost only upon rejection when re-prediction is necessary.
As shown in~\cref{tab:re_ptp}, using self-speculative decoding, PTP achieves significantly higher acceptance rates on OCR tasks and maintains superior rates on VLU and LM tasks.

\begin{figure*}[!t]
\centering
\includegraphics[scale=0.85]{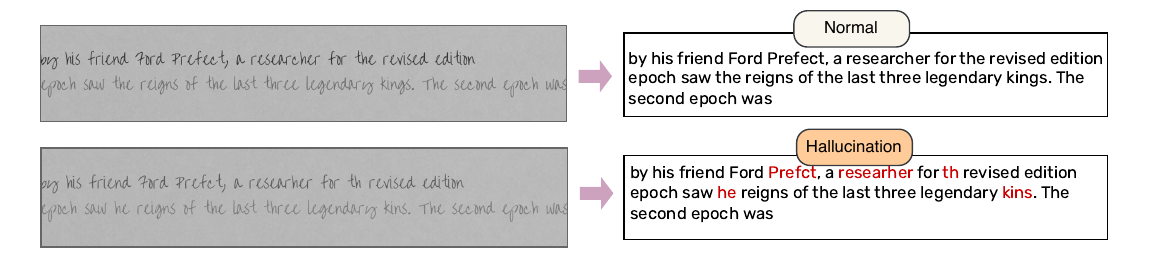}
\caption{Sampled rendered images and ground truth texts from normal and hallucinated datasets.}
\label{fig:hal_case}
\end{figure*}

\section{Details of Experiments}

\subsection{Detailed Evaluation Metrics}

In this section, we elaborate on the evaluation metrics employed in our experiments, which can be categorized into two groups: metrics for assessing model accuracy and metrics for measuring inference efficiency.

The following are the metrics used to evaluate the model's performance:
\begin{itemize}
    \item \textbf{Edit Distance:} we use the Levenshtein distance to measure the minimum number of single-character operations (insertions, deletions, or substitutions) required to transform the prediction into ground truth.
    
    \item \textbf{Character Detection Matching (CDM):} CDM is proposed by \cite{wang2025image} and is used to evaluate formula recognition performance, which renders both the model-predicted LaTeX and the ground-truth LaTeX formulas into image-formatted formulas, then employs visual feature extraction and localization techniques for precise character-level matching, incorporating spatial position information
    
    \item \textbf{Acceptance Rate:} Accpet rate is an evaluation metric in speculative decoding, which is the percentage of draft tokens accepted during verification. 
    In PTP, we treat future tokens predicted by register tokens as draft tokens and compute their acceptance rate during verification.
    
\end{itemize}

The following are the metrics used to evaluate the model's efficiency:
\begin{itemize}
    \item  \textbf{Latency:} The time from sending the request to receiving the final token on the user end, which directly affects perceived responsiveness.
    
    \item \textbf{Time Per Output Token (TPOT):} TPOT measures the average time required to generate each output token during the decoding stage of inference. The average time gap between generating each output token during the decoding stage of inference. A lower TPOT means the model can produce tokens faster, leading to higher tokens per second.
        
    \item \textbf{Inter-Token Latency (ITL):} The exact pause between two consecutive tokens.
    
    \item  \textbf{Throughput:} Throughput describes how much work an LLM can do within a given period. In this pepr, we focus on Output Tokens per Second (TPS), which provides a finer-grained view of throughput by measuring how many tokens are processed every second across all active requests.
\end{itemize}

\subsection{Analysis of the Number of Register Tokens}
\begin{figure}[!h]
    \centering
    \includegraphics[width=0.45\textwidth]{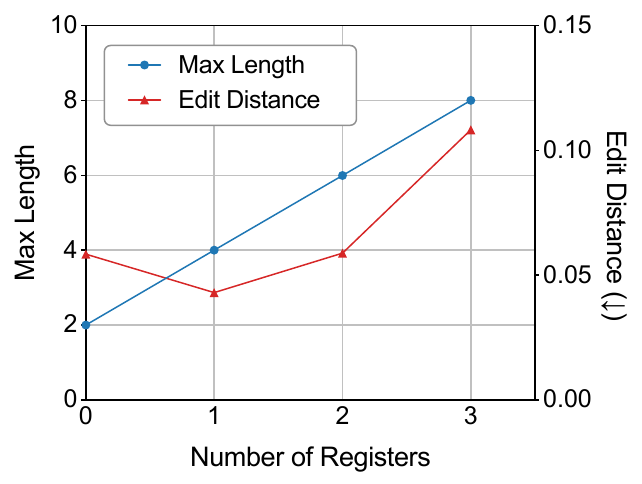}
    \caption{Analysis of the umber of register tokens.}
    \label{fig:num_reg}
\end{figure}

In this section, we investigative the effect of the register tokens numbers during training and inference phases.
While register tokens enable future token prediction, indiscriminately increasing their number is counterproductive.
As illustrated in \cref{fig:num_reg}, increasing register tokens proportionally expands the input sequence length during training, incurring significant computational overhead. Furthermore, excessively distant predictions suffer from error propagation and increased prediction complexity, ultimately degrading model accuracy. 
In our experiments, we set the number of register tokens to 2, which achieves an optimal trade-off between inference efficiency and model accuracy.

\begin{figure*}[!t]
\centering
\includegraphics[scale=0.75]{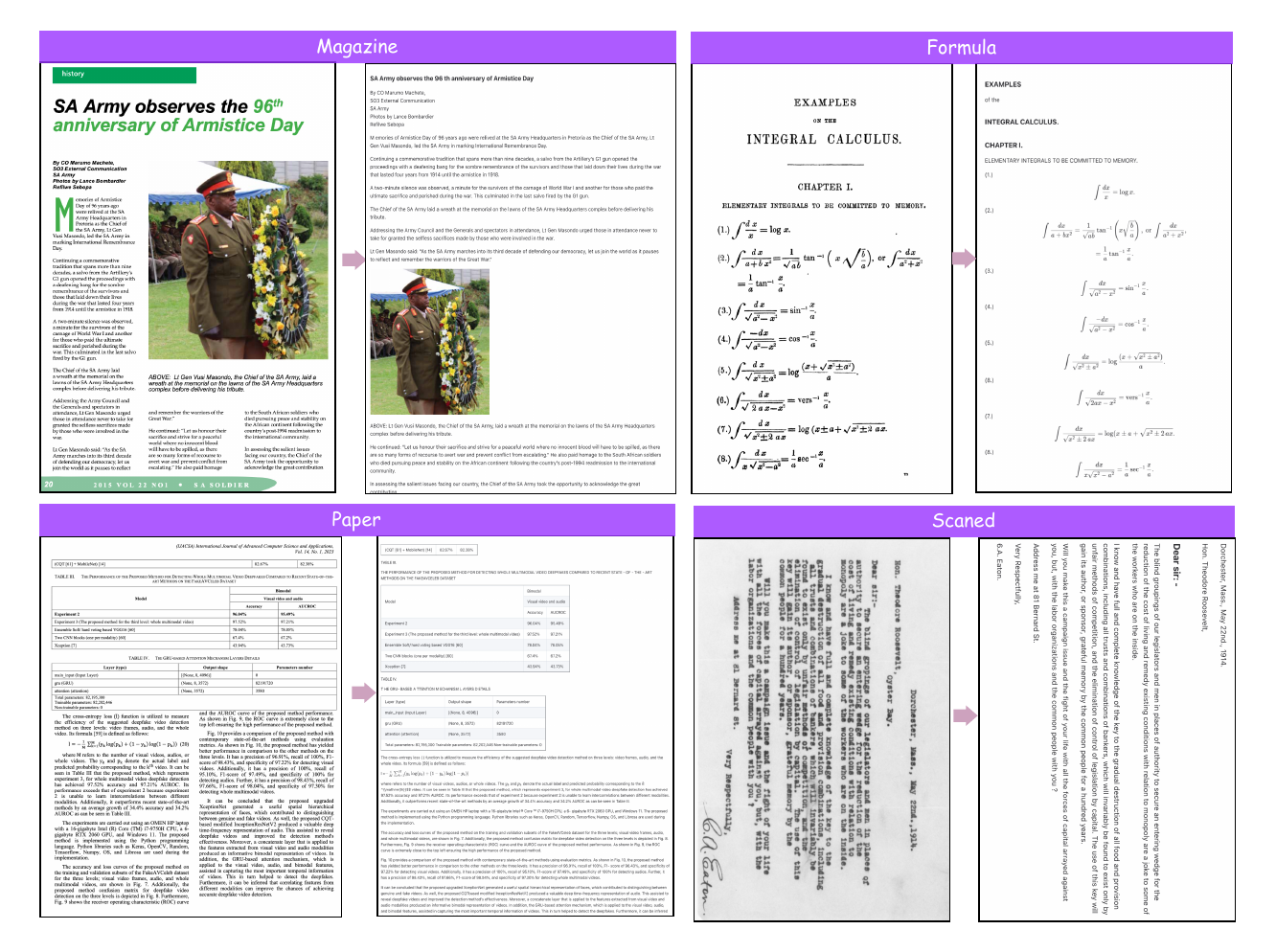}
\caption{The rendered markdown output for various types of documents of olmOCR-bench.}
\label{fig:case}
\end{figure*}

\subsection{Details of Hallucination Evaluation}
As illustrated in \cref{fig:hal_case}, we construct a OCR hallucination test set by randomly deleting or substituting characters in words to interfere with the VLM. We ensure consistency in background color, font, and rendering style between normal and hallucination test sets, with variations limited to text content and corresponding labels. 
As shown in \cref{fig:analysis} (Middle), PTP consistently outperforms NTP on both normal and hallucination data, while demonstrating superior robustness to hallucination-inducing perturbations, thereby validating its effectiveness in hallucination mitigation.

\section{Qualitative Examples}

We present several qualitative examples in~\ref{fig:case}.

\end{document}